# Assessing the value of a candidate
## Comparing belief function and possibility theories

**Didier Dubois**
Institut de Recherche
en Informatique de Toulouse
Université Paul Sabatier,
118 route de Narbonne
31062 Toulouse Cedex 4, France

**Michel Grabisch**
Thomson-CSF,
Corporate Research
Laboratory
Domaine de Corbeville
91404 Orsay, France

**Henri Prade**
Institut de Recherche
en Informatique de Toulouse
Université Paul Sabatier,
118 route de Narbonne
31062 Toulouse Cedex 4, France

**Philippe Smets**
IRIDIA
Université Libre
de Bruxelles
50 av. Roosevelt, CP 194-6
1050 Bruxelles, Belgium

**Abstract**

The problem of assessing the value of a candidate is viewed here as a multiple combination problem. On the one hand a candidate can be evaluated according to different criteria, and on the other hand several experts are supposed to assess the value of candidates according to each criterion. Criteria are not equally important, experts are not equally competent or reliable. Moreover levels of satisfaction of criteria, or levels of confidence are only assumed to take their values in linearly ordered scales, whose nature is often qualitative. The problem is discussed within two frameworks, the transferable belief model and the qualitative possibility theory. They respectively offer a quantitative and a qualitative setting for handling the problem, providing thus a way to compare the nature of the underlying assumptions.

## 1 Introduction

The problem of assessing the value of a candidate (it may be a person, an object, or any abstract entity for instance) is often encountered in practice. It is usually a preliminary step before making a choice. Such a problem can be handled in different manners depending on what the evaluation is based. One may have expert generic rules which aim at classifying candidates in different categories (e.g., 'excellent','good',... ,'very bad'). One may have a base of cases made of previous evaluations from which a similarity-based evaluation of the new candidate can be performed. This corresponds to two popular approaches in Artificial Intelligence (namely, expert systems and case-based reasoning), that one might also like to combine. In the following, the value assessment problem is rather posed in terms of multiple criteria, whose value for a given candidate can be more or less precisely assessed with some level of confidence by various experts (whose opinions are also to be fused). An important issue in such a problem is that the assessement of the value of a criteria for a candidate by an expert and their relevance for the selection itself are often pervaded with uncertainty and imprecision. Numerical and symbolic analysis of the data can be considered. In this paper, we show how the transferable belief model (TBM, [Smets, 1998]) and the qualitative possibility theory (QPT) can be applied to the selection problem, thus illustrating a numerical and a symbolic approach. Besides showing how the methods can be applied, their comparison indicates what are the underlying assumptions, what can help the potential user in selecting a method.

The paper is organized in four main parts. The problem is first precisely stated and the questions that it raises are pointed out. Then the two proposed approaches are presented and illustrated on the same example. Lastly, a comparative discussion of the two approaches takes place.

## 2 The multiple expert multiple criteria assessment problem

The value of a candidate for a given position has to be assessed by a decision maker (DM). For this evaluation, $m$ criteria are used. For each criterion, the value of the candidate is assessed (may be imprecisely) by $n$ experts (or sources). Criteria can be rank-ordered according to their importance for the open position. Their relation to the global "goodness" score of the candidate for the position may be also available. Each expert provides a precise or imprecise evaluation for each criterion and a level of confidence is attached to each evaluation by the expert who is responsible of it. The general reliability of each expert is also qualitatively assessed by DM.

Some general comments about the problem which is so far informally stated, have to be made. With a set of candidates, one may want to i) choose the best one(s); ii) rank all candidates from best to worst; iii) give a partial order between them, with possibly some incomparabilities; iv) cluster the candidates in several groups (e.g., good ones, bad ones, those with a weak point w.r.t. one criterion etc.). In fact, many methods in decision analysis proceed by a pairwise comparison of candidates which supposes to know all of them from the beginning. In this paper, we looks for a global evaluation of a candidate which may be unique. In case the problem would be to rank-order candidates only, a lexicographic approach (based on the comparaison of vectors made of scores of each candidate w.r.t. all criteria ordered according to their importance) might be enough. However such an approach assumes that a complete order exists between scores (which is not in agreement with the fact that the scores may be imprecise and pervaded with uncertainty).

The overall assumption of the model underlying the example is that there exist true scores for each individual criteria and for a global goodness score, all attached to the candidate, the last being a function of the first ones. The problem is that these true values are unknown to the DM and must be estimated from the expert opinions.

We now introduce some notations. The candidate is denoted $K$, when necessary. Criteria are numbered by $i$: $i = 1, \ldots, m$. The true (unknown) value of the level of satisfaction of criterion $i$ for the candidate $K$ is denoted $c_i(K)$, with $c_i(K) \in L_s$, where $L_s$ is the ordinal scale of levels of satisfaction, e.g. $L_s = \{1, 2, 3, 4, 5\}$ with 1=very bad,..., 5=very good. An element of $L_s$ is denoted $s$.

The evaluation of each criterion for $K$ and a given expert $j$ may be imprecise (either due to the fact that it is unclear to what precise extent $K$ satisfies criterion $i$ or due to the possible lack of competence in $i$ of the expert $j$ assessing the value). Each evaluation will be represented by a possibility distribution (discounted in case of limited expertise) restricting the more or less possible values of this evaluation. Let $\pi^j_{c_i(K)}$ ($\pi^j_i$ for short) the possibility distribution restricting the possible values of $c_i(K)$ according to expert $j$.

Experts are numbered by $j$: $j = 1, \ldots, n$. The function $\pi^j_i$ is a mapping from $L_s$ to $L_\pi$. In the example we shall use $L_\pi = \{\emptyset, a, b, \mathbb{1}\}$ which is an ordinal scale, where $\emptyset$ corresponds to impossibility, and $\mathbb{1}$ to total possibility. The true (unknown) global score of $K$ is denoted $c(K) \in L_s$. The associated possibility distribution is $\pi_{c(K)}$, or simply $\pi$. It represents the final result of the assessment procedure. Again $\pi : L_s \mapsto L_\pi$. The confidence level of expert $j$ in his assessment when judging criterion $i$ is denoted $\gamma_{ij}$, and these levels are defined on an ordinal scale $L_\gamma$ which can be related to $L_\pi$ as we shall see. The confidence of the decision maker into expert $j$'s opinions are denoted $\alpha_j$, $j = 1, \ldots, n$, and the $\alpha_j$'s are defined on an ordinal scale $L_\alpha$. For instance, $L_\alpha = \{\emptyset, r, s, \mathbb{1}\}$ with $\emptyset$ = not confident at all, $r$ = not very confident, ..., $\mathbb{1}$ = very confident. The levels of importance of criteria are denoted $\beta_i$, $i = 1, \ldots, m$, and the $\beta_i$'s also belong to an ordinal scale $L_\beta$. For instance $L_\beta = \{\emptyset, e, f, g, \mathbb{1}\}$ with $\emptyset$ = not important at all, $e$ = not very important, ..., $\mathbb{1}$ = very important.

In the following, $\vee$ and $\wedge$ denote max and min on a given ordinal scale. $\neg x$ for any $x$ in a given ordinal scale $L = \{\emptyset, s_1, \ldots, s_k, \mathbb{1}\}$ denotes the value corresponding to the reversed scale (i.e. $\neg \emptyset = \mathbb{1}, \neg s_i = s_{k-i+1}$). It is the counterpart to $1 - x$ on the $[0, 1]$ interval scale. For simplifying the notations we have denoted the top and bottom elements of each scale by the same symbols; this does not mean that there are the same however.

## 3 Data of the example

For illustrative purpose we use the following example with $m = 6$ criteria and $n = 4$ experts. Imagine that in some company, a new collaborator $K$ has to be hired for the marketing department. Six criteria are used for assessing his qualifications: analysis capacities (Ana), learning capacities (Lear), past experience (Exp), communication skills (Com), decision-making capacities (Dec) and creativity (Crea). The four experts are the directors of Marketing (Mkt), Financial (Fin), Production (Prod), and Human Resources (HR) departments. The data are summarized in Table 1.

On one side this example is intentionaly complex as we want to illustrate the power and flexibility of the two approaches. On the other side, the way the criteria should be aggregated is unspecified. In real life, all the subtleties introduced here are not necessarily present at the same time, which may make simpler.

## 4 The TBM approach

To represent belief functions, we use the next convention: $bel_Y^\Omega[PEV](\omega_0 \in A)$ denotes the strength of the weighted opinion (called belief) held by the agent $Y$ about the fact that the actual world $\omega_0$ belongs to $A$, a subset of the frame of discernment $\Omega$, given the piece of evidence $PEV$. The same indexing is used for basic belief assignments and plausibility functions induced by $bel$, with the symbol $bel$ replaced by $m$ or $pl$.

Many beliefs are used in this example. They are analogous to subjective probabilities, except they do not satisfy the additivity rule of probability theory. Their operational meaning and their assessment are obtained by methods similar to those used by the Bayesians [Smets and Kennes, 1994].

| $C_{ij}$ | Mkt D | Fin D | Prod D | HR D |
|---|---|---|---|---|
| Ana  |       | 4     | 2     |       |
| Lear | [2,3] | [1,5] | 4     | [2,4] |
| Exp  | 4     |       |       |       |
| Com  | 4     |       |       | 4     |
| Dec  | [1,5] | [1,5] | [1,2] | 3     |
| Crea | 5     |       |       | 1     |

| $\gamma_{ij}$ | Mkt D | Fin D | Prod D | HR D | $\beta_i$ |
|---|---|---|---|---|---|
| Ana  | $\emptyset$ | $\mathbb{1}$ | $\mathbb{1}$ | $\emptyset$ | $g$ |
| Lear | $b$ | $a$ | $b$ | $\mathbb{1}$ | $e$ |
| Exp  | $\mathbb{1}$ | $\emptyset$ | $\emptyset$ | $\emptyset$ | $e$ |
| Com  | $\mathbb{1}$ | $\emptyset$ | $\emptyset$ | $\mathbb{1}$ | $\mathbb{1}$ |
| Dec  | $a$ | $a$ | $a$ | $a$ | $g$ |
| Crea | $\mathbb{1}$ | $\emptyset$ | $\emptyset$ | $\mathbb{1}$ | $\mathbb{1}$ |
| $\alpha_j$ | $\mathbb{1}$ | $r$ | $r$ | $s$ | |

Table 1: Assessment by each director on each criteria of candidate K using $L_s = 1, 2, 3, 4, 5$ (above). Self-confidence of directors w.r.t. criteria (below).

In this section, we consider only candidate $K$ and all beliefs are held by one DM, so we can avoid mentioning them. Let $c_i$ be the actual (but unknown) value of the level of satisfaction of criterion $i$. Let $C_{ij}$ denote the data collected from director $j$ on criteria $i$ (for K), $C_{ij}$ being the intervals given in Table 1. Missing data are equated to the whole interval [1, 5].

Let $\Pi(C_{ij} = [a,a]|c_i = x)$ be the degree of possibility that Director $j$ states $C_{ij} = [a,a]$ with $a \in \{1, ..., 5\}$ given $c_i = x$ with $x \in \{1, ..., 5\}$. It represents the link between what the director will say and the actual value $c_i$. For simplicity sake, we use the same possibility function for every criterion and every director. For each criterion, we assume that the possibility is 1 when $a = c_i$, .5 when $|a - c_i| = 1$, and 0 otherwise. More complex possibility functions could be used, the proposed one being only "reasonable". It reflects the idea that it is (fully) possible that the director states the actual value, 'quite possible' that he states a value at one level deviation from the actual value, and impossible at two level deviations (director can be wrong, but the error will be small).

The degree of plausibility held by DM (before considering DM's opinions about Director $j$'s opinions) that the actual value of $c_i$ is in $A \subseteq L_{s_i}$ where $L_{s_i}$ is the domain of $c_i$ (equal to $L_s$), given what the Director $j$ states about criteria $i$, is given by:

$$pl^{L_{\cdot i}}[C_{ij}](c_i \in A) = \Pi(c_i \in A|C_{ij})$$
$$= max_{x \in A} \Pi(c_i = x|C_{ij})$$
$$= max_{x \in A} \Pi(C_{ij}|c_i = x)$$
$$= max_{x \in A, a \in C_{ij}} \Pi([a,a]|c_i = x)$$

It results from the classical relations $pl(A) = \Pi(A)$, $\Pi(X|Y)\Pi(Y) = \Pi(Y|X)\Pi(X)$, and $\Pi(X) = max_{x \in X} \Pi(x)$. We also assume a total a priori ignorance on $C_{ij}$ and $c_i$, i.e., $\Pi(C_{ij}) = 1$ for all $C_{ij}$ and $\Pi(c_i) = 1$ for all $c_i$.

The DM has some weighted opinions about the reliability of the assessment provided by Director $j$ on criteria $i$, represented by the coefficients $\gamma_{ij}$ (rescaled into $g_{ij} = 0, 1, 2, 3$ for $\gamma_{ij} = \emptyset, a, b, \mathbb{1}$, respectively). DM also has some prior opinions about the importance that he should give to Director $j$'s opinions when it comes to evaluating a candidate for a position like the open one, opinions that were coded by $\alpha_j$ (rescaled into $s_j = 1, 2, 3$ for $\alpha_j = r, s, \mathbb{1}$, respectively). All these opinions are transformed into discounting factors that express how much belief DM should give to the beliefs induced by the data produced by Director $j$ on criteria $i$. The discounting factors $d_{ij}$ are decreasing when the $\alpha$ and $\gamma$ values increase.

Discounting factors are meta-beliefs, i.e., beliefs over beliefs. They were introduced in [Shafer, 1976], and their formal nature explained in [Smets, 1993a]. Their real assessment is done as for any belief. For the purpose of the example, we use the values obtained from the relation:

$$d_{ij} = 1 - .95 \times (g_{ij}/3) \times (.75 + .25 \times (s_j - 1)/3).$$

Given the coefficients $d_{ij}$, DM discounts $bel^{L_{\cdot i}}[C_{ij}]$ into $bel^{L_{\cdot i}}[E_{ij}]$ where:

$$m^{L_{\cdot i}}[E_{ij}](A) = (1 - d_{ij}) \times m^{L_{\cdot i}}[C_{ij}](A) \text{ if } A \neq [1,5],$$

$$m^{L_{\cdot i}}[E_{ij}]([1,5]) = (1 - d_{ij}) \times m^{L_{\cdot i}}[C_{ij}]([1,5]) + d_{ij}$$

The belief function $bel^{L_{\cdot i}}[E_{ij}]$ represents DM's beliefs about the actual value of $c_i$ given what Director $j$ has stated and DM's opinions about $j$'s opinions (what is denoted by $E_{ij}$).

Then DM combines the Directors' opinions over Criterion $i$ by applying Dempster's rule of combination to the discounted belief functions. The resulting belief function represents DM's beliefs about the actual value of $c_i$.

$$bel^{L_{\cdot i}}[\&_j E_{ij}] = \oplus_j bel^{L_{\cdot i}}[E_{ij}]$$

The real problem for DM is not to assess the actual value of each criterion, but to assess if candidate K is 'good' for the position to be filled. So we introduce a 'Goodness Score', that will vary from 1 to 5, 1 for 'very bad', 5 for 'very good'. The relation between the Goodness Score and the value of criterion $i$ depends only on $\beta_i$, the level of importance of criterion $i$. The values of $f_i$ are tabulated in Table 2. E.g., when $\beta_i = g$, DM accepts as 'good' (score 5) a candidate for whom the criterion value is 4 or 5. smaller than 5. In practice, these values must be assessed through an analysis of the compensatory relation between the scores. E.g., it is as 'good' to have a score 3 for a criterion which importance is 1 than to have a score 3 when the importance is $g$ and to have a score 2 when the importance is $e$.

| $f_i(x)$ | 1 | 2 | 3 | 4 | 5 |
|---|---|---|---|---|---|
| $e$ | 1 | 3 | 4 | 5 | 5 |
| $f$ | 1 | 2 | 4 | 5 | 5 |
| $g$ | 1 | 2 | 3 | 5 | 5 |
| 1 | 1 | 2 | 3 | 4 | 5 |

Table 2: Values of goodness score $f_i(x)$ given the score $x$ of criterion $i$ (from 1 to 5) and its importance $\beta_i$

Let $f_i(x)$ be the value of the Goodness Score when the score for Criterion $i$ is $x$ and let $f_i(A) = \{f_i(x) : x \in A\}$. Then the belief over the value of the criterion $i$ is transformed into a belief $bel^G[C_i]$ over the Goodness Score by:

$$m^G[C_i](B) = \sum_{A: f_i(A)=B} m^{L_{s_i}}[\&_j E_{ij}](A)$$

$$\text{for } B \subseteq [1,2,3,4,5]$$

This just means that the basic belief mass given to A is transferred to the image of A under the transformation $f_i$ that holds between the criterion value and the Goodness Score.

The belief function $bel^G[C_i]$ represents DM's beliefs about the Goodness of the candidate given what DM collected about the criteria $i$. These belief functions are then combined on $i$ by Dempster's rule of combination. The resulting belief function represents DM's beliefs over the Goodness Score of candidate K given the collected information.

When it comes to decide which candidate to select, this last belief function is then transformed into a (pignistic) probability, denoted $BetP^G$, by the pignistic transformation (the nature and the justification of which are given in [Smets and Kennes, 1994]). This probability function over the actual value of the Goodness Score can then be used to order various candidates, using the classical methods developed in probability theory for such an ordering.

For the example under analysis, the basic belief masses $m^G$ for the Goodness Score and the pignistic probabilities $BetP^G$ are given in Table 3. In conclusion, there is a strong support that this candidate is Good. An expected Goodness Score can be computed, which value is 4.61, and it could be used to compare K with other candidates.

| focal sets | 4 | 5 | {4,5} |
|---|---|---|---|
| $m^G$ | .34 | .58 | .08 |
| $BetP^G$ | .38 | .62 | |

Table 3: Focal sets of $m^G$ and their masses $m^G$ for the Goodness Score, and non zero values of $BetP^G$ on the singletons of G.

## 5 The QPT approach

The problem raises three main questions: i) the representation of the precise or imprecise score of the candidate provided by each expert for each criterion, including the expert's confidence in his assessment; ii) the fusion of expert opinions; iii) the multiple criteria aggregation. The first step is easily handled in the possibilistic framework.

The possibility distribution of the true value of each score on criterion $i$ according to expert $j$, taking into account his competence, is computed in the following way. Interval-valued scores (including single values) are modelled by a possibility distribution taking the value 1 in the interval and 0 outside. Blanks (absence of answers) are interpreted as a possibility distribution being 1 everywhere (modelling "unknown"). The confidence level $\gamma_{ij}$ is taken into account by a discounting process, defined as ($\tilde{\pi}$ denotes the original possibility distribution function ($\pi$.d.f.))

$$\pi_i^j(s) = \tilde{\pi}_i^j(s) \vee \neg\gamma_{ij}, \quad \forall s \in L_s \quad (1)$$

Note that the certainty level $\gamma_{ij} \in L_\gamma$ is turned into a possibility level $\neg\gamma_{ij}$ over the scores not compatible with $\tilde{\pi}_i^j$. Thus $L_\gamma$ is $L_\pi$ reversed (usual equivalence between certainty of $A$ and possibility of not $A$). This leads to Table 4. In each cell the $\pi$.d.f is enumerated on $L_s$.

| | Mkt D | Fin D | Prod D | HR D |
|---|---|---|---|---|
| Ana | 11111 | 00010 | 01000 | 11111 |
| Lear | a11aa | 11111 | aaa1a | 01110 |
| Exp | 00010 | 11111 | 11111 | 11111 |
| Com | 00010 | 11111 | 11111 | 00010 |
| Dec | 11111 | 11111 | 11bbb | bb1bb |
| Crea | 00001 | 11111 | 11111 | 10000 |

Table 4: Opinions of directors on each criterion

As already said, the global assessment requires two types of combination: a *multicriteria aggregation* prob-lem, and the *fusion* of expert evaluations. So, depending on the way the problem is presented, we may either think of i)first computing the global evaluation of $K$ according to each expert and then to fuse these evaluations into a unique one, or ii)on the contrary, first fuse the expert evaluations for each criterion, and then aggregate the "global" results pertaining to each criterion. In general, the two procedures are not equivalent.

At this point, it is worth emphasizing that expert opinion fusion and multicriteria aggregation are two operations which do not convey the same intended semantics. The fusion of expert opinions aims at finding out what are the possible values of the genuine score of $K$ for a given criterion, and possibly to detect conflicts between experts. Hopefully, some consensus should be reached at least on values which are excluded as possible values of the score. The aggregation of multiple criteria evaluations aims at assessing the global worth of the candidate from his scores on the different criteria; then different aggregation attitudes may be considered, e.g., conjunctive ones where each criterion is viewed as a constraint to satisfy to some extent, or compensatory ones where trade-offs are allowed.

In the following, we choose to merge expert's opinions on each criterion first, and then to perform a multicriteria aggregation, since it might seem more natural to use the experts first to properly assess the score according to each criterion. Proceeding in the other way would assume that each expert is looking for a global evaluation of the candidate (may be using his own criteria aggregation attitude) and the decision maker is only there for combining and weighting expert's evaluations. Since we first merge the $\pi$.d.f's, one of the allowable methods is to use a weighted maximum. Indeed disjunctive combination is advisable in case of conflicting opinions [Dubois and Prade, 1994], as it is the case here (conjunctive fusion can be applied only if there is no conflict). A disjunctive combination means that the opinion of an important expert will be taken into account, even if this can conflict with another important one. However, unreliable estimates should be discounted in the disjunction. The reliability $w_{ij}$ attached to $\pi_i^j$ should be both upper bounded by the confidence $\alpha_j$ of DM in the expert $j$ and the self-confidence $\gamma_{ij}$ of the expert. This leads to take the weight $w_{ij}$ as the conjunctive-like combination of $\alpha_j$ and $\gamma_{ij}$, $\forall i = 1, \ldots, 6$. The weighted disjunction is defined as:

$$\pi_i'(s) = \bigvee_j \left[ w_{ij} \wedge \pi_i^j(s) \right], \quad \forall s \in L_s, \quad (2)$$

As said above, $w_{ij} = \gamma_{ij} \otimes \alpha_j$ where $\otimes$ is a conjunction operator from $L_\gamma \times L_\alpha$ to $L_\gamma$. Table 5 defines $\otimes$ on the basis of an implicit commensurateness hypothesis of $L_\gamma$ and $L_\alpha$.

| $\otimes$ | $\emptyset$ | $r$ | $s$ | $\mathbb{1}$ |
|---|---|---|---|---|
| $\emptyset$ | $\emptyset$ | $\emptyset$ | $\emptyset$ | $\emptyset$ |
| $a$ | $\emptyset$ | $a$ | $a$ | $a$ |
| $b$ | $\emptyset$ | $a$ | $b$ | $b$ |
| $\mathbb{1}$ | $\emptyset$ | $a$ | $b$ | $\mathbb{1}$ |

Table 5: Definition of $\otimes$

| $w_{ij}$ | Mkt D | Fin D | Prod D | HR D |
|---|---|---|---|---|
| Ana | $\emptyset$ | $a$ | $a$ | $\emptyset$ |
| Lear | $b$ | $a$ | $a$ | $b$ |
| Exp | $\mathbb{1}$ | $\emptyset$ | $\emptyset$ | $\emptyset$ |
| Com | $\mathbb{1}$ | $\emptyset$ | $\emptyset$ | $b$ |
| Dec | $a$ | $a$ | $a$ | $a$ |
| Crea | $\mathbb{1}$ | $\emptyset$ | $\emptyset$ | $b$ |

Table 6: Weights $w_{ij}$ for all criteria and directors

Table 6 gives the weights for all criteria and experts. We apply formula (2) to $\pi$.d.f's of Table 4, and obtain the $\pi$.d.f's of Table 5 (left).

We notice that some distributions are no longer normalized (i.e., no score is fully possible at level $\mathbb{1}$). This is because the weights are not normalized, i.e. $\vee_j w_{ij} < \mathbb{1}$ for some $i$'s, which means that the DM cannot be fully confident in any director when assessing some criteria (namely Ana, Lear and Dec).

We should either use the unnormalized distributions to fully keep track of the problem, or normalize them in a suitable way. Here, we choose the second solution, and propose the following approach. We consider that the maximum of the distribution, denoted $h$, reflects the uncertainty level, considered to be $\neg h$ in the information. This means that the amount of conflict is changed into a level of uncertainty and the minimum level of the modified distributions will be $\neg h$. Then, we make an additional hypothesis that the scale is a interval scale (which is questionable!) so that the profile of the distribution is not changed. Specifically:

$$\pi_i(s) = \pi_i'(s) + \neg(\vee_s \pi_i'(s)) \quad (3)$$

It means that + in (3) is defined by $s_i + s_j = s_{\min(k+1, i+j)}$ on a scale $\{s_0 = \emptyset, s_1, \ldots, s_k, s_{k+1} = \mathbb{1}\}$. The result is shown in Table 5. A slightly different method would be to keep possibility degrees equal to $\emptyset$ at $\emptyset$, since $\pi_i'(s) = \emptyset$ means that all experts agree on the fact that the value $s$ is impossible.

| Ana | $\emptyset a \emptyset a \emptyset$ | | Ana | $b\mathbb{1}b\mathbb{1}b$ |
|---|---|---|---|---|
| Lear | $abbba$ | | Lear | $b\mathbb{1}\mathbb{1}\mathbb{1}b$ |
| Exp | $000\mathbb{1}0$ | | Exp | $000\mathbb{1}0$ |
| Com | $000\mathbb{1}0$ | | Com | $000\mathbb{1}0$ |
| Dec | $aaaaa$ | | Dec | $\mathbb{1}\mathbb{1}\mathbb{1}\mathbb{1}\mathbb{1}$ |
| Crea | $b000\mathbb{1}$ | | Crea | $b000\mathbb{1}$ |

Table 7: Merged opinions of directors on criteria (left); Merged opinions after normalization (right)

The way of aggregating the criteria evaluations is not at all specified in the statement of the problem. Only qualitative levels of importance $\beta_i$ are provided for each criterion $i$. Even with ordinal scales, different attitudes can be thought of. The aggregation may be purely conjunctive (based on "min" operation), or somewhat compensatory (using a median operation for instance). It might be also disjunctive: at least one important criteron is to be satisfied, it is modelled by a weighted maximum. More general aggregation attitudes can be captured by Sugeno's integral ([Sugeno, 1977];[Grabisch et al., 1995]). Let us assume that the DM has a conjunctive attitude.

We use the weighted minimum method in what follows for the aggregation. With our notations, in the case of precise scores, the global score is obtained by

$$c = \bigwedge_{i=1}^{6} [(\neg \beta_i) \tilde{\vee} c_i] \quad (4)$$

where $\tilde{\vee}$ is a disjunctive operator from $L_\beta \times L_s$ to $L_s$. Table 8 gives the definition of this operator.

| $\tilde{\vee}$ | $\emptyset$ | $e$ | $f$ | $g$ | $\mathbb{1}$ |
|---|---|---|---|---|---|
| 1 | 1 | 2 | 3 | 4 | 5 |
| 2 | 2 | 2 | 3 | 4 | 5 |
| 3 | 3 | 3 | 3 | 4 | 5 |
| 4 | 4 | 4 | 4 | 4 | 5 |
| 5 | 5 | 5 | 5 | 5 | 5 |

Table 8: Definition of $\tilde{\vee}$

The "fuzzy" evaluation provided by the $\pi$.d.f.'s can be handled in different ways. A natural manner to proceed is to extend multicriteria aggregation techniques to such non-scalar evaluations (this can be done both if this step is done before or after the fusion step). At the technical level, this will be done using an extension principle [Zadeh, 1975] which enables to extend any function/operation $f$ to any fuzzy arguments. Namely, $f(P_1, \ldots, P_m)(s) = \vee_{s=f(s^1, \ldots, s^m)} P_1(s^1) \wedge \ldots \wedge P_m(s^m)$ where $P_i(s^i)$ is the possibility degree of score $s^i$ according to $\pi$.d.f. $P_i$. Here the result will be a possibility distribution $\pi_K$ restricting the possible values of the global evaluation of $K$. Then a fuzzy ranking method should be used for assessing to what extent it is certain, and to what extent it is possible that $K_1$ is better than $K_2$, on the basis of $\pi_{K_1}$ and $\pi_{K_2}$ in case of several candidates ([Dubois and Prade, 1983]).

Since scores of criteria are imprecise, the extension principle has to be used. We have for any $s \in L_s$

$$\pi(s) = \bigvee_{s=\bigwedge_{i=1}^{6}[(\neg \beta_i) \tilde{\vee} c_i]} (\pi_1(c_1) \wedge \cdots \wedge \pi_6(c_6)) \quad (5)$$

Applying this formula to our data, we finally obtain the distribution given in Table 9. The result can be

| score | 1 | 2 | 3 | 4 | 5 |
|---|---|---|---|---|---|
| possibility degree | $b$ | $\mathbb{1}$ | $\mathbb{1}$ | $\mathbb{1}$ | 0 |

Table 9: Possibility distribution $\pi$ of the final score interpreted by saying that the candidate $K$ is certainly not a very good candidate, but $K$ is most likely between bad and good. Let us briefly comment the result. The most important criteria are Com and Crea, where the best scores are respectively 4 and 5 with possibility $\mathbb{1}$. This explains why the global value 5 is impossible, since it is not reached by one important criterion. Besides, there is a conflict on Crea since score 1 has possibility degree $b$, which would lead to a global score equal to 1. Now observe that the score on criterion Dec which is rather important, is completely unknown, and may take any value between 1 and 5. This explains why intermediate values 2, 3 are also possible for the global score in a minimum-based aggregation.

Note that if weighted min combinations are used both in the expert opinion fusion and in the multi-criteria aggregation then the two combinations commute.

## 6 General discussion

The reader may be astonished by the discrepancy between the results obtained by the belief function approach and by the possibilistic approach. This does not reflect any disagreement between the two approaches, but in fact two different options in the aggregation mode of the criteria. Indeed, the TBM approach has used a compensatory aggregation mode, while weighted minimum has been used in the QPT approach (although it would be possible to use median operators in the qualitative setting in order to allow for compensation). The min-based combination is the only conjunctive idempotent attitude, whereas Dempster's rule of combination is not idempotent (even though there exist other combination rules that are more 'cautious' and idempotent).

Moreover, when introducing the Goodness Score in the belief function approach, the underlying idea is that we are all the less demanding for reaching the maximum level of satisfaction as the criterion is less important. This is not at all the understanding chosen in the possibilistic approach (remember that the problem is not completely specified). Rather, the unimportant criteria were considered as somewhat satisfied even if they were not at all satisfied.

In fact there are different possible ways of understanding the weighting of the criteria, in a qualitative conjunctive setting. Let $c_i(K)$ be the supposedly precise value of score of $K$, according to criterion $i$. Indeed, we can i) modify $c_i(K)$ into $\neg \beta_i \tilde{\vee} c_i(K)$ as we did, or

ii) modify $c_i(K)$ into $\begin{cases} 5 & \text{if } c_i(K) \geq \beta_i \\ c_i(K) & \text{if } c_i(K) < \beta_i \end{cases}$ ($\beta_i$ is interpreted as a threshold level to reach to be fully satisfied, taking advantage of the commensurateness hypothesis underlying Table 8). Choosing the option (ii) above in the QPT approach instead of (i) would be similar to the idea of Goodness Score as defined in the TBM approach.

Another difference is that the possibility distribution we have obtained represents the uncertainty pervading the genuine value of the candidate. It is not a scalar expectation, a thing which can be also computed in the possibilistic framework as explained now.

The assessment problem can be viewed as a *decision* to be made *under uncertainty*. Namely, the relevance of the criteria defines a candidate profile, i.e., a kind of utility function, while the value of the candidate $K$ is ill-known with respect to each criterion (once expert opinions have been fused). Viewing the problem in this way, it is natural to compute to what extent it is certain (or it is possible) that the candidate $K$ (whose level assessment may be pervaded with imprecision and uncertainty for each criteria) satisfies the criteria at the required level; see ([Dubois and Prade, 1995]) for an axiomatic view of the corresponding qualitative decision procedure. It corresponds to a fuzzy pattern matching problem ([Dubois et al., 1988]) in practice.

For each criterion $i$, we build a satisfaction profile $\mu_i$, e.g., $\mu_i(s) = \neg \beta_i \vee s$. $\mu_i$ means that the greater the score, the better the candidate, and that the satisfaction degree is lower bounded by $\neg \beta_i$ which is all the greater as $i$ is less important. Then the certainty that $K$ satisfies the profile is given by

$$\wedge_i \wedge_s (\mu_i(s) \vee \neg \pi_{c_i(K)}(s)) \qquad (6)$$

where $\pi_{c_iK}$ is supposed to be normalized. The possibility that $K$ satisfies the profile is

$$\wedge_i \vee_s (\mu_i(s) \wedge \pi_{c_i(K)}(s)) \qquad (7)$$

where $\pi_{c_i(K)}$ has been obtained by fusing the expert opinions first, at the level of each criterion. The possibility degree (very optimistic) should be only used for breaking ties in case of equality of the certainty degrees for different candidates. The aggregation by $\wedge_i$ of the elementary certainty and possibility degrees can be justified in the possibility framework (definition of a join possibility distribution of non-interactive variables ([Zadeh, 1975]), and interpretation of the global requirement in terms of a weighted conjunction of elementary requirements pertaining to each criterion). The expression (6) and (7) can be viewed as possibilistic "lower and upper expectations".

What is computed remains in the spirit of the approach detailed before. Instead of obtaining a pos-sibility distribution we obtain two scalar evaluations for which it should be possible to show that they summarize this possibility distribution (in a sense to be made precise at the theoretical level). Anyway both approaches first compute the $\pi_{c_i(K)}$'s and are based on the choice of a weighted conjunction.

Lastly, it is also important to judge the (potential) explanation capabilities. Are the results provided by the system easy to explain to the user if necessary? E.g., why the value of the candidate is finally assessed as such? Such explanation capabilities are presented in [Farreny and Prade, 1992] and in [Xu and Smets, 1996] for what concerns the QPT and the TBM, respectively.

For summarizing the main differences between the belief function (TBM) and the possibility-based approaches (QPT) on the value assessment problem, it is clear that the possibilistic method can handle poor data expressed in a qualitative, non-numerical, way whereas the belief function framework may more easily capture reinforcements and compensatory effects.

## 7    Concluding remarks

In this paper, we have taken advantage of a generic value assessment problem for discussing the different facets of the problem and raising the various difficulties and hypotheses which should be made at each step for computing a meaningful evaluation through two models, the QPT and the TBM.

The purpose of the paper was twofold: first identifying and discussing various facets of a value assessment problem, and second showing how two different approaches (which have in common the capability of modelling imprecision) can handle the problem in manners which are in fact quite parallel.

The statement of the problem contains no number, but only ordinal assessments. Owing to the qualitative framework of possibility theory , the QPT provides an obvious solution. However note that it is important to consider the nature of the scales which are used in order to know what operations are meaningful on them. Moreover commensurateness hypotheses are necessary.

With the TBM, numbers (the beliefs) are needed. They are in fact analogous to those a probability approach would ask for. The difference between the TBM and a probability approach is that the TBM accepts and uses the data just as they are, without introducing extraneous data. E.g., a probability approach would allocate (equal) probabilities to each of the individual values of the scores when the score is only known as a non-degenerated interval.

Sensitivity analysis can be performed, where either the values of the parameters (TBM) or the aggregation operators (QPT) are slightly modified and the robustness of the conclusions can be assessed. Furthermore the DM can also determine the sensitivity of the results if more precise data were collected,. Given this information, the DM could efficiently ask to a specific expert to provide a better estimate of the candidate score for a specific criterion, avoiding getting "useless" data.

One important point in practice for this type of problem is also to be able to provide *explanations* to the user on the results. The qualitative framework of possibility theory allows for a logical reading and processing of the evaluations. This offers a potential for explanation capabilities.

## 8  Acknowledgements

This work has been supported by the European working group FUSION (Esprit Project 20001: Fusion of Uncertain Data from Sensors, Information Systems and Expert Opinions). Special thanks are due to Simon Golstein who helps stating the candidate value assessment problem considered in this paper.